\pdfoutput=1

\documentclass[10pt, a4paper]{article}

\usepackage{lrec-coling2024} %

\usepackage{booktabs}
\usepackage{amsmath}
\usepackage{lscape}

\title{ParaNames 1.0: Creating an Entity Name Corpus for 400+ Languages using Wikidata*\thanks{* This is a preprint version of this article. Please cite the version published in the \href{https://aclanthology.org/}{LREC-COLING 2024 proceedings in the ACL Anthology}.}}

\name{Jonne S{\"a}lev{\"a} and Constantine Lignos} 

\address{Michtom School of Computer Science \\
  Brandeis University\\
  \texttt{\{jonnesaleva,lignos\}@brandeis.edu} \\}

\abstract{
We introduce ParaNames, a massively multilingual parallel name resource consisting of 140 million names spanning over 400 languages.
Names are provided for 16.8 million entities, and each entity is mapped from a complex type hierarchy to a standard type (PER/LOC/ORG).
Using Wikidata as a source, we create the largest resource of this type to date.
We describe our approach to filtering and standardizing the data to provide the best quality possible.
ParaNames is useful for multilingual language processing, both in defining tasks for name translation/transliteration and as supplementary data for tasks such as named entity recognition and linking.
We demonstrate the usefulness of ParaNames on two tasks. First, we perform canonical name translation between English and 17 other languages. Second, we use it as a gazetteer for multilingual named entity recognition, obtaining performance improvements on all 10 languages evaluated.
 \\ \newline 
}

\begin{document}

\maketitleabstract

\section{Introduction}

Our goal in creating ParaNames is to introduce a massively multilingual entity name resource that provides names for diverse types of entities in the largest possible set of languages and can be kept up to date through a nearly automated preprocessing procedure.
A large resource of names of this type can support development and improvement of multilingual language technology applications, as it is often important to know how real-world entities are represented across various languages.

The correspondences of names across languages are not always easy to model; they can involve a mix of transliteration and translation and often involve inconsistencies across languages or even among names in a given language.
As a concrete example, some country names are translated in Finnish, so \emph{United Kingdom} is written as \emph{Yhdistynyt kuningaskunta}, a literal, word-by-word, translation.
In contrast, smaller territories may or may not be translated: the U.S. states of \emph{North Carolina} and \emph{New York} are written as \emph{Pohjois-Carolina} (with \emph{North} translated) and \emph{New York}, respectively. 
Moreover, Finnish versions of the U.S. states are often idiosyncratically translated, e.g. California is represented as \emph{Kalifornia}, whereas Colorado is represented as \emph{Colorado}. 
The examples above demonstrate the complex choices that language speakers make in representing named entities---even when only dealing with Latin script---and underscore the need for a large-scale, multilingual resources of entity name correspondences to effectively model these phenomena.

Most research groups (ours included) lack the means to assemble annotators in hundreds of languages to produce a carefully, manually curated resource with the coverage we desire.
Even with sufficient means, such a resource would quickly fall out of date and would be difficult to grow over time.

Instead, our approach is to try to adapt an existing, continuously maintained data source to serve this purpose.
This presents several requirements:
first, the data source itself needs to cover as broad a set of languages as possible, especially under-resourced ones.
Second, the process of constructing the resource needs to be near-automatic to allow updates as the upstream data source is modified.
Finally, to have the most useful set of names in each language possible, we need to try to exercise proper quality control, for example ensuring that the entities in each language are in the desired script even when there are errors in the source data.

We selected \href{https://www.wikidata.org}{Wikidata} as our data source due to its nature as a perpetually updating collection that enables continuous improvement and expansion, as well as its extensive coverage of languages and entities.
In this paper, we present our approach to transforming the Wikidata knowledge graph into a dataset of person, location, and organization entities with parallel names.
We identify potential problems in the source data---such as the lack of standardization of the script(s) used in each language---and provide a processing pipeline that addresses them.
In addition to ensuring consistency in the scripts used for each language, we focus on making the names as parallel as possible by removing extraneous information that can accompany them, such as parenthetical explanations.
We hope to provide regular updates to this resource to include corrections and improvements to both Wikidata and our extraction process.
ParaNames is licensed using the Creative Commons Attribution 4.0 International (CC BY 4.0) license and was first publicly released in March 2021.\footnote{ParaNames is released at \url{https://github.com/bltlab/paranames}. Earlier versions of this work were presented in non-archival forms at \href{https://arxiv.org/abs/2104.00558}{AfricaNLP 2021} and \href{https://aclanthology.org/2022.sigtyp-1.15/}{SIGTYP 2022}.}

\section{Related Work}
\label{sec:relatedwork}

While there is previous work in the construction of multilingual name resources, we are not aware of an \emph{openly-accessible} resource containing the names of millions of \emph{modern} entities.

\citetlanguageresource{wu-etal-2018-creating} create a translation matrix of 1,129 biblical names, with each English name containing translations into up to 591 languages.
\citet{merhav-ash-2018-design} release bilingual name dictionaries for English and each of Russian, Hebrew, Arabic, and Japanese Katakana.
However, their resource is limited to a few languages and only covers single token person names.

The Named Entity Workshop (NEWS) shared task has created parallel name resources across a series of shared tasks.
In the 2018 version of the shared task \citep{chen-etal-2018-report,chen-etal-2018-news}, participants were asked to transliterate between language pairs involving English, Thai, Persian, Chinese, Vietnamese, Hindi, Tamil, Kannada, Bangla, Hebrew, Japanese (Katakana / Kanji), and Korean (Hangul), although the task did not include transliteration between all pairs.
The NEWS 2018 datasets are hand-crafted and much smaller than ours, at most 30k names per language pair.
The datasets for these shared tasks are not fully publicly available; the test set is held back, and each of the five training sets is subject to different licensing restrictions.

There is scattered prior work on extracting parallel names from Wikipedia and Wikidata.
One of the earliest explorations of extracting parallel names from Wikidata at scale was performed by \citetlanguageresource{irvine-etal-2010-transliterating}.
\citetlanguageresource{steinberger-etal-2011-jrc} also collected names for roughly 200,000 entities in 20 scripts and several languages, using Wikipedia and news articles as their data sources.
Building on their work, \citetlanguageresource{benites-etal-2020-translit} also used Wikipedia as a data source and automatically extracted potential transliteration pairs, combining their outputs with several previously published corpora into an aggregate corpus of 1.6 million names.
While all these works produced collections of entities that are more modern than those produced by e.g. \citetlanguageresource{wu-etal-2018-creating}, the total number of names is still far smaller than ParaNames.

One line of research has focused on evaluating the quality of Wikidata-derived labels. For instance, \citet{StatisticalAndAmaral2022} showed that the ``also-known-as'' metadata may contain more useful labels than the main labels that Wikidata provides, and that sentence embeddings can be used to pick the best label for each pair in a way that improves the degree to which the names are parallel. However, their experimental results focus on a manually annotated dataset of 10 languages, which allowed them to do much more manual annotation and quality control than in our work.

Specifically for lower-resourced languages, many approaches to named entity recognition and linking for the LORELEI program \citep{strassel-tracey-2016-lorelei} used Wikidata, Wikipedia, DBpedia, GeoNames, and other resources to provide name lists and other information relevant to the languages and regions for which systems were developed.
However, while ad-hoc extractions of these resources were integrated into systems, we are unable to identify prior attempts to create a transparent, replicable extraction pipeline and to distribute the extracted resources with wide language coverage.

\section{Dataset Construction}
\label{sec:data-extraction-and-quality-challenges}

Each entity in Wikidata is associated with several types of metadata, including a set of one or more names that different languages use to refer to it.
To construct ParaNames, we began by extracting all entity records from Wikidata and ingesting them into a MongoDB instance for fast processing.
Given that we are working with such a large-scale dataset, there are quality control decisions that arise when working with the data, which we describe in this section.

\subsection{Harvesting Entity Names}

The most important type of metadata for us are the various names for each entity.
In Wikidata, each entity includes one \textit{main label} (denoted by \texttt{rdfs:label} in RDF dumps) and zero or more secondary labels/aliases (denoted by \texttt{skos:altLabel} in RDF dumps and by ``also-known-as'' on Wikidata). 
Even though these aliases contain useful examples of real-world names for many entities, after manual inspection we opted to only use the main labels in ParaNames.

This decision was based on the observation that the aliases often include names that only loosely correspond to the canonical name of the entity.
For example, AKAs for the late U.S. Supreme Court justice Ruth Bader Ginsburg (\href{https://www.wikidata.org/wiki/Q11116}{Q11116}) contain not only her full name, \emph{Ruth Joan Bader Ginsburg}, but also common aliases from popular culture, such as \emph{Notorious RBG}.  
In the case of Donald Trump (\href{https://www.wikidata.org/wiki/Q22686}{Q22686}) the AKAs contain other variations of his name (\emph{Donald John Trump}, \emph{Donald J. Trump}, etc.), but also pseudonyms that he has used that do not correspond to his actual name (\emph{John Barron}, \emph{John Miller}, \emph{David Dennison}, etc.).
While this information could be argued to be useful for downstream tasks such as entity linking, we felt that these alternative names introduced potentially unwanted variation in the names across languages.
For this reason, we chose not to include the also-known-as fields in our dataset at this time.

\subsection{Language Representation}

The number of entities for each language varies wildly across Wikidata. In ParaNames, we exclude languages with only a single name as having a single name would not constitute meaningful representation of the language.
The number of languages that entities have labels in varies as well.
For example, the entry for Alan Turing (\href{https://www.wikidata.org/wiki/Q7251}{Q7251}) shows his name written in over a hundred languages, including many that use non-Latin scripts.
Internally, each language is referred to using a Wikimedia language code, which may or may not correspond one-to-one with natural languages.
Often there are several Wikimedia codes for a given spoken language, varying in script or geography.\footnote{
The relationship between Wikimedia language codes and other language codes is rather complex. Originally, the Wikimedia language codes were designed to comply with \href{https://datatracker.ietf.org/doc/html/rfc3066}{RFC3066}, but there are inconsistencies and \href{https://meta.wikimedia.org/wiki/Wiki_language_ISO_639-1_\%E2\%86\%92_BCP_47_proposal}{standardization is unlikely to occur soon}.
Some, but not all, of the language codes are identical to modern  \href{https://datatracker.ietf.org/doc/html/rfc5646}{BCP 47 codes (RFC5646)}.
In this paper, we try to distinguish between the Wikimedia language codes---which may identify a language along with a script, geographical region, or dialect---and higher-level language identifiers which use only the first two letters of the language code.
}

For example, the Kazakh language is associated with the Wikimedia language codes \texttt{kk} (Kazakh), \texttt{kk-arab} (Kazakh in Arabic script), and \texttt{kk-latn} (Kazakh in Latin script).
At other times, the language codes are specific to geography rather than writing system. In the case of Kazakh, there are three main geography-specific language codes: \texttt{kk-cn} (Kazakh in China), \texttt{kk-kz} (Kazakh in Kazakhstan) and \texttt{kk-tr} (Kazakh in Turkey).
In our analysis and the resource we distribute, we do \textit{not} combine these, as the more detailed language codes may be helpful in learning to transliterate between different scripts of the same language.
Our code for reproducing ParaNames includes an option to toggle this behavior by setting \texttt{should\_collapse\_languages=\{yes, no\}} as desired.\footnote{We do normalize language codes to match ISO-639-3 language codes in select special cases: for Bhojpuri (\texttt{bho/bh}), we normalize all the names to the language code \texttt{bho}. We also normalize Cantonese (\texttt{zh-yue/yue-hant/yue}) to \texttt{yue}, Hokkien/Southern Min (\texttt{zh-min-nan/nan}) to \texttt{nan} and Samogitian (\texttt{sgs/bat-smg}) to \texttt{sgs}.}

\subsection{Script Usage and Standardization}

While language codes can identify a specific script for a language, many Wikidata labels do not conform to the scripts expected for each language.
While this may be a data quality issue in some languages, the presence of several scripts can also reflect real world-usage depending on the language.
For example, Kazakh uses both the Cyrillic and Arabic alphabets, thus multiple scripts are to be expected across a collection of names and ParaNames reflects this diversity.

We chose to standardize the names within for each language by filtering out names that are not in the desired script(s) for the language.
An example of this would be a Russian entity label like \emph{Canada} which is not written in Cyrillic.
While we explored automated methods of doing this, ultimately we decided that manually constructing a list of allowed scripts for each language would yield the best results.
For each language, we used Wikipedia as an authoritative source to look up which scripts are used to write the language and filtered out all names that were not primarily written in one of the allowed scripts for that language. For each name, we used the PyICU library\footnote{\url{https://gitlab.pyicu.org/main/pyicu}}  to identify the Unicode script property of each character in the name and chose the most frequent script as the primary script for the entire name.

To quantify how much this filtering changed the entity names associated with each language, we attempted to measure script uniformity for each language.
For each language, we aggregated the Unicode script tags produced by PyICU across names for each language and computed the entropy of this distribution, calling this quantity \emph{script entropy} and used it as a proxy for script consistency within a language's names.
Languages whose names are consistently written in a single script will have near-zero entropy.
The filtering process decreased the average script entropy from 0.16 to 0.03.
After filtering, 487 language codes remained with a total of 140,178,539 names across 16,834,537 entities.

\subsection{Providing Entity Types}
\label{sec:providing-entity-types}

Even though entities often have detailed information about what they represent, Wikidata does not directly categorize entities as instances of higher-level types such as location (LOC), organization (ORG), and person (PER).
To obtain this information, we extracted entity types based on the Wikidata inheritance hierarchy.
Specifically, we identified suitable high-level Wikidata IDs---Q5 (human) for PER, Q82794 (geographic region) for LOC, and Q43229 (organization) for ORG---and classified each Wikidata entity that is an instance of these IDs as the corresponding named entity type.

While the \texttt{instance-of} relation is transitive---all instances of a subtype are instances of the higher-level type---we noticed that taking all subtypes of these high-level types led to many entities that were not individual persons to be classified as PER, such as \textit{Government secretaries of Policies for Women of the State of Bahia} (\href{https://www.wikidata.org/wiki/Q98414232}{Q98414232}).
To exclude such entities, we required that PER entities must also explicitly be an instance of Q5 (person) in addition to any subclass types.
As we did not observe similar problems for LOC and ORG entities, we kept the typing rules unchanged for them.
If we had imposed a more stringent type requirement as we did for PER, it would have dramatically decreased the number of entities for LOC and ORG.

As shown in Table~\ref{tab:entitytypes}, a relatively small number of entities get assigned to multiple types. While this is a result of multiple-inheritance in the entity type hierarchy of Wikidata, having multiple types is not incorrect as an entity can represent several different types. 
In ParaNames, we opted to preserve this information, as assigning only a single type to complex entities could make our dataset less useful by ignoring inherent entity typing uncertainty.

\begin{table}[tb]
\small
\centering
\begin{tabular}{lrr}
\toprule
\textbf{Entity type} &      \textbf{Count} & \textbf{Percentage} \\
\midrule
PER   &  10,002,138 &  59.41\% \\
LOC   &   3,880,088 &  23.05\% \\
ORG   &   2,631,350 &  15.63\% \\
Mixed &     320,961 &   $<$2\% \\
\midrule
\textbf{Total}           & 16,834,537 &    100.0\% \\
\bottomrule
\end{tabular}

\caption{Number of entities and percentage of all entities assigned to each combination of LOC, ORG and PER in ParaNames.}
\label{tab:entitytypes}
\end{table}

\subsection{Matching Information Across Languages}

Ideally, the names in ParaNames would be maximally parallel so that both names in a name pair contain the same information.
However, some Wikidata names contain additional information in parentheses, intended to help disambiguate between similar-looking entities.
For instance, the entity with the English label \emph{Wang Lina (boxer)} (\href{https://www.wikidata.org/wiki/Q60834172}{Q60834172}) has a Russian label which contains the translation of the word \emph{boxer} in parentheses.
However, this is not the case for all languages: for example, the Spanish name for the entity is simply \emph{Wang Lina}.
To standardize the amount of information per name across languages, we remove all parentheses and tokens inside them using a regular expression.

\section{Canonical Name Translation}

\begin{table}[tb]
\small
    \centering
    \begin{tabular}{llrr}
    \toprule
      Language &                    Script &  Names & \% Train \\
    \midrule
        Arabic &                    Arabic & 500,000 &      11.1\% \\
      Japanese &         Kanji, Kana$^{*}$ & 500,000 &      11.1\% \\
      Swedish &                      Latin & 500,000 &      11.1\% \\
      Russian &                   Cyrillic & 500,000 &      11.1\% \\
      Persian &                     Arabic & 457,200 &      10.2\% \\
    Vietnamese &                     Latin & 429,185 &       9.6\% \\
    Lithuanian &                     Latin & 282,074 &       6.3\% \\
        Hebrew &                    Hebrew & 205,704 &       4.6\% \\
        Korean &                    Hangul & 203,042 &       4.5\% \\
      Latvian &                      Latin & 177,577 &       4.0\% \\
      Armenian &                  Armenian & 161,957 &       3.6\% \\
         Greek &                     Greek & 149,515 &       3.3\% \\
        Kazakh &                  Cyrillic & 124,574 &       2.8\% \\
          Urdu &                    Arabic & 103,803 &       2.3\% \\
          Thai &                      Thai &  72,112 &       1.6\% \\
      Georgian &                  Georgian &  70,965 &       1.6\% \\
         Tajik &          Cyrillic,  Latin &  52,574 &       1.2\% \\
    \midrule
    Total &                                &         &     100.0\% \\
    \bottomrule
    \end{tabular}
    \caption{Parallel training data statistics and the script(s) used to write the names in our dataset. The development and test sets were each balanced to 5,000 names per language. $^{*}$Kana jointly refers to the two Japanese syllabaries, Hiragana and Katakana.}
    \label{tab:parallel_data_stats}
\end{table}

We demonstrate the applicability of ParaNames by using it to train models on two downstream tasks: canonical name translation and named entity recognition.
As our first task, we use the parallel names in ParaNames to translate entity names from many languages to English and from English to many languages.
We call this task \emph{canonical name translation}, as the task is to translate the Wikidata label (canonical name) for an entity into the label in another language.

It is important to clarify what this task is and what it is not.
We do not refer to this task as name transliteration because not every name pair is strictly transliterated; often the mapping includes elements of transliteration, translation (especially for organization names), and sometimes morphological inflection/deinflection as well.
The task is also not the translation of a name within a sentence, which often requires correct morphological inflection of the name in its sentential context.

\subsection{Data Selection and Splitting}
    
For our experiments, we translate named entities from 17 languages---Arabic, Armenian, Georgian, Greek, Hebrew, Japanese, Kazakh, Korean, Latvian, Lithuanian, Persian (Farsi), Russian, Swedish, Tajik, Thai, Vietnamese, and Urdu---into English and vice versa, using a single multilingual model for each translation direction.
We chose these languages as they cover a wide geographic distribution, as well as several different language families, typological features, and orthographic systems.

When evaluating our model's performance on Latin-script languages, it is important to avoid inflating performance numbers by having a large part of the evaluation set consist of names written identically to English. 
To accomplish this, we include a small number of such languages in our experiments---Swedish, Vietnamese, Lithuanian, Latvian, and Tajik. 
The languages we selected also have varying amounts of data available in our corpus.
All the languages we selected had sufficient names to allow for the development and test sets to be equally balanced across languages (5k names per language), but there was an order of magnitude difference between the language with the fewest names available for the training data (Tajik, ~50k). 
We limited those languages (Arabic, Japanese, Russian, Swedish) to 500k names in training to avoid oversampling.

To create the parallel data for this task, we extracted all Wikidata IDs that had names in English and at least one of the other languages in our selected set.
We divided the IDs into either the train, development, or test set using an 80/10/10 split.
The overall statistics of the parallel data can be seen in Table~\ref{tab:parallel_data_stats}.
While per-ID splitting does not guarantee identical language stratification across train, development, and test sets, we employ it to avoid a data leakage scenario where the English side of a given entity name might appear in more than one of our train, development, or test sets.
Notably, this leakage does occur in the data split created by \citetlanguageresource{wu-etal-2018-creating} because they split the data by source-target name pairs, not entities.

To further balance our datasets and avoid overly biasing our models towards the higher-resourced languages, we also capped the maximum number of examples per language at 500,000 for the training data and 5,000 for the development and test data.

\paragraph{Special tokens}
After creating the data splits, we augment the source side of each name pair with one or more ``special tokens'' that provide information about the non-English language\footnote{In the case of X $\rightarrow$ En models, this corresponds to the source language and in En $\rightarrow$ X models the target language.} and the underlying entity. 
The purpose of the special token(s) is to help our model better separate languages, especially ones with potentially overlapping scripts such as Tajik, Russian and Kazakh or Swedish, Latvian, and Lithuanian.

Specifically, we map each source-side string S into an augmented version based on the regular expression\footnote{The notation \texttt{p?} refers to zero or one occurrences of a pattern \texttt{p}.}
$$
S \mapsto (\text{<L>})\text{?} (\text{<S>})\text{?}(\text{<T>})\text{? } S
$$

\noindent where $\text{<L>}$ refers to the non-English language of the name pair, $\text{<S>}$ refers to the script the non-English name is written in, and $\text{<T>}$ refers to its assigned entity type.
As a concrete example, when performing name translation from Swedish to English, the Swedish entity name \textit{Hyde Park} (\href{https://www.wikidata.org/wiki/Q123738}{Q123738}) would first be tokenized at the character level into \textit{H y d e  P a r k} and could then be augmented as e.g.\textit{<sv> <Latin> <LOC> H y d e  P a r k} depending on the experiment.
We outline the special token combinations we experiment with in Section~\ref{sec:tag-ablation-experiments}.

\subsection{Model Details}

The model we use is a simple character-level Transformer-based translation model trained from scratch on the ParaNames data.
We use the model structure and hyperparameters from past transliteration experiments by \citet{moran-lignos-2020-effective} with minor changes.
We use a 4-layer Transformer with a hidden layer size of 1024, embedding dimension of 200, 8 attention heads, and a learning rate of 0.0003, with a dropout probability of 0.2. The label smoothing parameter is set to 0.1, and batch size is set to 128.
We use the Adam optimizer for a maximum of 75,000 updates. Each experiment is repeated 5 times using random seeds ranging from 1917 to 1921. 
A single NVIDIA RTX 3090 GPU was used for both training and decoding.
For each experimental condition (i.e. direction, source-side special token setting, and language), training the model took roughly 9 hours and evaluation took roughly 15–30 minutes.
We implemented our model using fairseq \citep{ott-etal-2019-fairseq}.

\subsection{Evaluation Metrics}
We evaluated using three metrics: 1-best accuracy (where each a name translation must match the reference \emph{exactly}), \emph{character error rate} (CER), computed analogously to word error rate but at the character level, and \emph{mean F1-score} based on longest common subsequence \citep{chen-etal-2018-news}, similar to ROUGE-L \citep{lin-2004-rouge}. 
For brevity, we limit the discussion here to accuracy, as the other metrics are largely correlated with it and the conclusions remain the same.
Full results with all metrics are reported in Table~\ref{bigtable-all-results} in the Appendix.

\begin{table*}[tb]
\small
\centering
\begin{tabular}{lrrrrrr}
\toprule
{} & \multicolumn{3}{c}{X $\rightarrow$ English} & \multicolumn{3}{c}{English $\rightarrow$ X} \\
    \cmidrule(lr){2-4} \cmidrule(lr){5-7}
Language &          Language only &                  Language + type & $\Delta$     &     Language only &           Language + type & $\Delta$ \\
\midrule
Georgian   &  31.56 $\pm$ 0.12 &  \textbf{32.67 $\pm$ 0.12} &       1.11 &  48.94 $\pm$ 0.16 &  \textbf{51.00 $\pm$ 0.23} &       2.06 \\
Arabic     &  30.77 $\pm$ 0.13 &  \textbf{31.88 $\pm$ 0.12} &       1.11 &  46.54 $\pm$ 0.09 &           46.77 $\pm$ 0.13 &       0.23 \\
Korean     &  31.21 $\pm$ 0.10 &  \textbf{32.28 $\pm$ 0.19} &       1.07 &  42.32 $\pm$ 0.10 &           42.57 $\pm$ 0.09 &       0.25 \\
Thai       &  38.55 $\pm$ 0.11 &  \textbf{39.59 $\pm$ 0.22} &       1.04 &  14.59 $\pm$ 0.13 &  \textbf{15.02 $\pm$ 0.04} &       0.43 \\
Persian    &  25.90 $\pm$ 0.12 &  \textbf{26.92 $\pm$ 0.12} &       1.02 &  41.90 $\pm$ 0.13 &           42.10 $\pm$ 0.21 &       0.20 \\
Urdu       &  26.18 $\pm$ 0.12 &  \textbf{27.02 $\pm$ 0.19} &       0.84 &  18.03 $\pm$ 0.20 &           17.96 $\pm$ 0.16 &      -0.07 \\
Hebrew     &  17.66 $\pm$ 0.12 &  \textbf{18.46 $\pm$ 0.15} &       0.80 &  37.32 $\pm$ 0.07 &  \textbf{37.83 $\pm$ 0.06} &       0.51 \\
Kazakh     &  48.36 $\pm$ 0.17 &  \textbf{49.14 $\pm$ 0.11} &       0.78 &  58.00 $\pm$ 0.11 &           58.30 $\pm$ 0.10 &       0.30 \\
Armenian   &  38.05 $\pm$ 0.09 &  \textbf{38.76 $\pm$ 0.16} &       0.71 &  47.42 $\pm$ 0.05 &  \textbf{47.95 $\pm$ 0.08} &       0.53 \\
Greek      &  31.04 $\pm$ 0.07 &  \textbf{31.67 $\pm$ 0.11} &       0.63 &  30.77 $\pm$ 0.07 &  \textbf{31.22 $\pm$ 0.15} &       0.45 \\
Lithuanian &  80.01 $\pm$ 0.07 &  \textbf{80.56 $\pm$ 0.08} &       0.55 &  78.73 $\pm$ 0.44 &           79.30 $\pm$ 0.31 &       0.57 \\
Russian    &  45.18 $\pm$ 0.04 &  \textbf{45.65 $\pm$ 0.06} &       0.47 &  42.42 $\pm$ 0.12 &  \textbf{43.26 $\pm$ 0.14} &       0.84 \\
Japanese   &  29.51 $\pm$ 0.09 &  \textbf{29.97 $\pm$ 0.13} &       0.46 &  26.95 $\pm$ 0.08 &           27.30 $\pm$ 0.09 &       0.35 \\
Latvian    &  74.86 $\pm$ 0.16 &           75.26 $\pm$ 0.03 &       0.40 &  73.05 $\pm$ 0.25 &  \textbf{73.81 $\pm$ 0.19} &       0.76 \\
Tajik      &  51.25 $\pm$ 0.15 &           51.56 $\pm$ 0.21 &       0.31 &  56.30 $\pm$ 0.10 &  \textbf{56.82 $\pm$ 0.11} &       0.52 \\
Vietnamese &  86.78 $\pm$ 0.07 &           87.02 $\pm$ 0.07 &       0.24 &  77.71 $\pm$ 0.28 &           78.17 $\pm$ 0.16 &       0.46 \\
Swedish    &  90.26 $\pm$ 0.12 &           90.34 $\pm$ 0.08 &       0.08 &  87.96 $\pm$ 0.20 &           88.31 $\pm$ 0.06 &       0.35 \\
\midrule
Micro-avg.     &  45.71 $\pm$ 0.06 &  \textbf{46.40 $\pm$ 0.05} &       0.69 &  48.76 $\pm$ 0.05 &  \textbf{49.27 $\pm$ 0.04} &       0.51 \\
\bottomrule
\end{tabular}
\caption{Canonical name translation accuracy when translating to and from English, sorted by descending performance difference ($\Delta$) on the X $\rightarrow$ En side.
Statistically significant differences from the language-only baseline are \textbf{bolded}.}
\label{acc-best-results}
\end{table*}

We first performed a baseline experiment using a source-side special prefix token that only conveys the language being translated into.
As our second set of experiments, we then modified the information contained in special tokens to assess the effects on performance.
The overall test set results for both translation directions are given in Table \ref{acc-best-results}.
The table reports the mean value and the standard deviation of the mean (standard error) computed across training five models with different random seeds.
We round all values to 2 decimal places.

\subsection{Baseline: Language Special Token}

As our first experiment, we evaluated canonical name translation performance in both X $\rightarrow$ En and En $\rightarrow$ X directions using language special tokens on the source side.
In terms of 1-best accuracy, our model exhibits an overall trend regardless of the translation direction: it performs best on Latin script-languages (Swedish, Vietnamese, Lithuanian, Latvian), followed by others roughly in proportion to how ambiguous the mapping from each language's script to the Latin alphabet is. 
For instance, in the X $\rightarrow$ English direction, the Latin-script languages are followed by those written in Cyrillic script (Tajik, Kazakh, Russian) which also makes sense as Cyrillic names can be transliterated into Latin script relatively unambiguously.
For the rest of the languages, the pattern is less clear and accuracy is roughly in the 25-50\% range.

Even though translation into English is generally easier for the model, switching the translation direction does not seem to substantially change the performance rankings for most languages.
A notable exception in the X $\rightarrow$ English direction is Hebrew which consistently ranks worst among all languages. 
This is most likely caused by the lack of vowels in the Hebrew names which the model must infer on the English side. When translating from English to Hebrew the accuracy improves substantially (37.32 vs 17.66), as the model does not have to infer the vowels, only delete them.
A similar but reversed pattern can be observed for Thai where accuracy is much higher when translating into English than when translating into Thai (14.59 vs 38.55). 
Since the Thai script indicates vowels using combining diacritics, we hypothesize this might be more difficult for the model to get exactly correct than English where vowels are written out explicitly.

Qualitatively, when inspecting model outputs, we noticed that often our model relies too heavily on transliteration when some words must be translated or vice versa.
Many outputs were also incorrect because they lacked extra information that was only present on the target side and omitted on the source side. 
For example, tokens like \textit{Stream} in \textit{Cuiva Stream} (\href{https://www.wikidata.org/wiki/Q21412684}{Q21412684}) are only present in the English name and cannot be learned by seeing the non-English source label.

\subsection{Finding the Optimal Special Tokens}
\label{sec:tag-ablation-experiments}

In addition to adding source-side language tokens to our parallel data, we also hypothesized that other information may also be relevant can be helpful in guiding the decoder.
For example, most person names are transliterated while organization names tend to include more translation, and many location name pairs contain tokens on one side that are absent from the other.
Script information can also be useful when dealing with languages that are written in several scripts or to help encourage transfer across languages that share a script.

To investigate these hypotheses, we repeated the first experiment using various different kinds of special tokens: a language token (\texttt{<Russian>}) in conjunction with either a type token (\texttt{<PER>}), a script token (\texttt{<Cyrillic>}), or both.
We also performed an ablation experiment by removing special tokens when possible.
Entity type tokens were generated from the PER/LOC/ORG type information in ParaNames inferred from Wikidata types.
For the small number of entities that mapped to multiple types, an arbitrary one was chosen.
Script tokens were generated using the PyICU library as with script filtering using the most frequent Unicode script in a particular name.

For the X $\rightarrow$ English direction, we experimented with the following special token configurations: no special token; script only; language only; language and script; language and entity type; language, entity type, and script. 
For English $\rightarrow$ X we only evaluated having a language token and language and entity type tokens, as fewer configurations were possible.
The language token must always be present for the model to know what language to translate into, so we did not experiment with removing it.
We could not use the script token for English $\rightarrow$ X as it is computed from the non-English (target) side of the translation; using it would effectively leak specific information about the test data as part of the model's job is to predict which script to use in the case of a language that uses multiple scripts.

The full results of our experiments across all languages, metrics and special token settings can be seen in Table~\ref{bigtable-all-results}.
For the X $\rightarrow$ English direction, using no special token or a script-only special token performs similarly to the baseline (not shown due to space limitations), but as shown in Table \ref{acc-best-results} there are clear improvements from adding an entity type special token---roughly 0.7 accuracy points micro-averaged over languages. 
Adding a script token to the language-only baseline or language + entity type setting seems to provide a marginal performance improvement, but the effect size seems ultimately consistent with random noise.
For the English $\rightarrow$ X direction, we can see that adding an entity type special token provides roughly a 0.5 point improvement.
In both directions, the setting with a language and entity type special token appears to perform best.

We  also performed statistical significance testing to assess whether there are differences between the various special token settings. 
For each language, metric, and translation direction, we performed a two-tailed Mann-Whitney U test, which is a non-parametric alternative to the two-sample $t$-test and requires no assumptions about the distribution of the data. 
For each test, we compared the baseline to our best special token setting with language and entity type tokens.
Our null hypothesis was that there is no difference between the medians of the two groups.
In Table~\ref{acc-best-results}, we use boldface to indicate where significant deviations from the language-only token baseline were observed and where a statistically significant result was obtained at the $p < 0.05$ level. The micro-averaged accuracy is significantly different from the baseline from in both translation directions. When translating to English, the null hypothesis is rejected for all languages with a raw difference from baseline above 0.4---this coincides exactly with the 13 languages with the largest difference from baseline. When translating from English, the pattern is the same, except for Lithuanian and Vietnamese for which we fail to reject the null hypothesis.

\section{Named Entity Recognition}

\begin{table*}[tb]
\small
\centering
\begin{tabular}{lrrrrrrrrl}
\toprule
  Language &  Tokens
   & Entities  & Links & Coverage (\%) & No gaz. &  Gazetteer &  $\Delta$ (diff.) &   $\sigma$ (SD) &  $\Delta/\sigma$ \\
\midrule
    Swahili &  79.2k &    2,074 &   533 &           25.70\% &         77.83 &      80.06 & \textbf{2.25} &        1.15 &     1.93* \\
    Finnish & 162.7k &    7,156 & 1,948 &           27.22\% &         63.04 &      65.18 & \textbf{2.14} &        1.57 &     1.37* \\
      Hausa &  80.2k &    2,124 &   296 &           13.94\% &         83.60 &      84.40 & \textbf{0.80} &        0.67 &     1.19* \\
     Yoruba &  83.2k &    2,073 &   219 &           10.56\% &         66.31 &      67.74 & \textbf{1.44} &        1.30 &     1.11* \\
       Igbo &  61.7k &    2,022 &   175 &           8.65\% &         79.22 &      79.75 & \textbf{0.54} &        1.03 &              0.52 \\
    Luganda &  46.6k &    2,721 &   145 &           5.33\% &         74.11 &      74.76 & \textbf{0.64} &        1.39 &     0.46* \\
      Wolof &  52.9k &      836 &    99 &           11.84\% &         59.12 &      59.58 & \textbf{0.45} &        1.99 &              0.23 \\
    Amharic &  37.0k &    2,037 &   125 &           6.14\% &         52.38 &      52.73 & \textbf{0.36} &        1.67 &              0.21 \\
      Hindi &   2.2M &   57,087 & 7,891 &           13.82\% &         92.07 &      92.09 & \textbf{0.02} &        0.14 &              0.11 \\
Kinyarwanda &  68.8k &    1,870 &    91 &           4.87\% &         63.02 &      63.14 & \textbf{0.12} &        1.32 &              0.09 \\
\midrule
     Median &           74k &           2,074 &          197 & 11.20\% &   70.21  &	71.25 &	1.04 &	1.31 &	0.80 \\
     Mean &           287.2k &          8,000  &          1,152 & 12.81\% &     70.18 &	71.05 &	0.87 &	1.22 &	0.71 \\
\bottomrule
\end{tabular}
\caption{
Micro-averaged F1 score on the MasakhaNER, HiNER  and TurkuNLP test sets. 
\textit{No gaz.} refers to our baseline model with no gazetteer features. \textit{Gazetteer} refers to the best soft gazetteer feature combination across configurations. 
$\sigma$. For individual languages, all positive $\Delta$ values are bolded; statistically significant $\Delta/\sigma$ values are marked with an asterisk.
}
\label{ner-table}
\end{table*}

As a second task, we focus on multilingual named entity recognition on a geographically diverse set of 10 languages.
We use ParaNames as a \textit{gazetteer}, seeking to improve entity typing performance with the extra information it contains. 
We experiment with 10 geographically diverse languages: Hindi, Finnish, Amharic, Hausa, Igbo, Kinyarwanda, Luganda, Swahili, Yoruba, and Wolof. 
For each language, we rely on recently published NER corpora to ensure we have quality training and evaluation data comparable to past work.
Specifically, for Hindi and Finnish, we use the HiNER \citeplanguageresource{murthy-etal-2022-hiner} and TurkuNLP \citeplanguageresource{luoma-etal-2020-broad} corpora; for all the African languages, we use the MasakhaNER corpus \citeplanguageresource{adelani-etal-2021-masakhaner}.

\subsection{Gazetteers}

To construct the gazetteers for a given language, we first extract all entity names in that languages from ParaNames.
As noted in Section~\ref{sec:providing-entity-types}, some entities contain multiple types. Whether to deduplicate them is not obvious without context, so we leave this open as a hyperparameter and train models using both a full and a deduplicated version of the gazetteer.
When deduplicating, we use the following disambiguation rules to ensure each Wikidata ID in our gazetteer is mapped to a single type: (ORG, PER) $\rightarrow$ ORG, (ORG, LOC) $\rightarrow$ LOC, (LOC, PER) $\rightarrow$ PER, and (LOC, ORG, PER) $\rightarrow$ ORG.

To link entity mentions in the NER data to entries in the gazetteer, we link each mention that exactly matches an entry in ParaNames. 
Following \citet{rijhwani-etal-2020-soft}, we reduce lookup times at train time by extracting all n-grams up to length 3 from our training data and linking those that exactly match an entity name in our gazetteer. This process is done entirely offline and does not add computational overhead at train time. 

\subsection{Model Details}

All our NER models use the CNN-BiLSTM-CRF architecture originally proposed by \citet{ma-hovy-2016-end}. 
We use the DyNet-based implementation by \citet{rijhwani-etal-2020-soft} and train our models for 50 epochs on a 64-core AMD Ryzen Threadripper CPU. 
As a baseline, we turn all gazetteer features off and use a plain CNN-BiLSTM-CRF model. 
As an alternative, we consider the model proposed by \citet{rijhwani-etal-2020-soft} which incorporates soft gazetteer features into the network at the LSTM and/or the CRF layer.
We consider the location of these gazetteer features---as well as the autoencoder loss used in the original paper---a hyperparameter and conduct a search over all 16 configurations.
Other hyperparameters are kept the same as in \citet{rijhwani-etal-2020-soft}, except for the word embedding and LSTM hidden state dimensions, which we also vary between the values 128, 200, and 256. The random seed is determined by a random integer.
The full set of hyperparameters can be seen in the Appendix in Table~\ref{tab:softgaz-ablation}.

\subsection{Evaluation Metrics}
We evaluate our models using standard exact-match entity F1 score, micro-averaged across types. 
Preliminary experiments showed that varying the dimensionality of the model resulted in about as much variance as changing the random seed for a given configuration. 
Thus, out of concern for ``fishing for noise,'' we did not pick the best-performing LSTM/word embedding dimension for each of the 16 configurations. %
Instead, we ranked the configurations based on their \textit{median} performance across the dimensionality grid and random seeds.

Our main results on the named entity recognition task are presented in Table~\ref{ner-table}. 
The \textit{No gaz.} and \textit{Gazetteer} columns show the micro-averaged F1 across all possible hyperparameter configurations and seeds.
Based on the last two rows, it is evident that using a gazetteer improves performance, with an absolute median improvement of 1.04 F1 points over the no-gazetteer baseline. The mean F1 improvement is 0.87 which is similar to the median, although slightly smaller due to the presence of ``outlier'' languages, e.g. Hindi, that have particularly low F1 differences.

Looking across languages, it appears that there is that there is a lot of variation in the magnitude of the absolute improvement over baseline ($\Delta$), ranging from 2.25 for Swahili to 0.02 for Hindi.
However, the within-language standard deviations ($\sigma$) also seem to vary substantially across languages: Wolof has the largest variation with a standard deviation of 1.99, whereas the standard deviation for Hindi is only 0.14.
This suggests that it is not meaningful to compare \textit{absolute} F1 differences from language to language as doing so would neglect relevant information about how much they vary simply due to random seeds or hyperparameters.

Instead, we divide the raw differences by the within-language standard deviation and obtain a normalized effect size, $\Delta$/$\sigma$.
Placing the values on a common scale (measured in units of standard deviations) also allows for better cross-linguistic comparisons. For example, for Igbo and Wolof, the raw F1 differences are fairly similar (0.54 vs. 0.45), but the standard deviations are quite different (1.03 vs. 1.99). 
If we normalize the effect sizes, we obtain 0.54/1.03 = 0.52 for Igbo and 0.45/1.99 = 0.23 for Wolof. 
These new numbers give a much clearer picture of where the model does well and properly reflect the difference in standard deviations.
The normalization also shrinks the mean and median effect sizes closer together: using $\Delta/\sigma$, the median and mean improvements over baseline are 0.80 and 0.71, respectively.

In terms of normalized effect sizes, our model produces effects above 1.0 on four out of our ten languages: Swahili, Finnish, Hausa and Yoruba. 
This can be interpreted as ParaNames giving a performance boost that is larger than random variation.
On other languages, the effect sizes are significantly smaller, ranging from 0.52 on Yoruba to 0.09 on Kinyarwanda. 
While thresholds are ultimately arbitrary, values of 0.5 and 0.8 have been suggested as thresholds for classifying effect sizes as ``medium'' and ``large'' \citep{sawilowsky2009new}. 
Using this interpretation, ParaNames produces ``medium-to-large'' performance improvements on 5 out of 10 languages (Swahili, Finnish, Hausa, Yoruba and Igbo).
Another way to assess whether using ParaNames as a gazetteer is helpful for a given language is to use statistical significance testing.
We perform another analysis in this vein and conduct a two-tailed Mann-Whitney U-test \citep{mannwhitney1947} for each language. In each case our null hypothesis is that both the baseline and gazetteer F1 scores come from the same distribution.
At the $\alpha < 0.05$ significance level our test rejects the null hypothesis for Swahili, Finnish, Hausa, Yoruba and Luganda which is in near-perfect agreement with the effect size-based interpretation.

Without multiple data sets per language, it is difficult to make predictions about what languages ParaNames would be most useful for across any possible dataset. Looking at Table~\ref{ner-table}, however, we can say that gazetteer coverage clearly seems to matter.
For the best-performing languages, several hundreds of mentions are linked to our gazetteer which represent over 10\% of all unique entities in the data. Swahili, for example has over 25\% coverage. Conversely, for the languages whose effect sizes are small, the data only contain very few linked mentions which also lowers any potential effect a gazetteer can have. For example, the model performs worst on Kinyarwanda, whose coverage is only 4.87\% with only 91 links in total. 

Wolof is a slight exception to this pattern, as its effect size is only 0.23 even though it has a slightly higher coverage ratio at 11.84\%. This can be explained by the low absolute number of linked mentions, 99.
Hindi, on the other hand, is a more notable exception: 7,892 out of 57,087 entities in the data (or 13.82\%) are linked to ParaNames, yet the effect size is minuscule at 0.11. 
A potential explanation for this is the fact that the Hindi NER data is several orders of magnitude larger than most of the languages, causing the model to have to rely less on outside information such as a gazetteer.

\section{Limitations}

\paragraph{One label per entity per language} ParaNames only uses the ``main label'' property in Wikidata to identify names for entities.
One of the limitations of this approach is that a given entity can only have a single label within a single Wikimedia language code, even though there may be multiple possible transliterations of an entity name.
This can be especially problematic for languages that use more than one script but for which a finer-grained language code the specifies the script, such as \texttt{sr-cyrl}, is not available.
For example, Bosnian only has the language code \texttt{bs} but is commonly written in Cyrillic and Latin scripts.
Wikidata does provide an ``also-known-as'' (AKA) property that may get around this limitation, but unfortunately, it often includes names that only loosely correspond to the canonical name of the entity such as the nickname \emph{Notorious RBG} for the late U.S. Supreme Court justice Ruth Bader Ginsburg (\href{https://www.wikidata.org/wiki/Q11116}{Q11116}).

ParaNames was extracted from Wikidata, and while we have undertaken significant efforts to clean and standardize the data, there will always be errors with a community-edited resource of this size.
Of particular concern is that some names may be blindly copied across languages, especially from English.
As we have filtered the name in each language by script, those languages that do not use Latin script may have higher-quality data, as it is not possible for names to have been copied into those languages from English without review.

\section{Conclusion}

ParaNames enables the cross-lingual modeling of names for millions of entities in over 400 languages.
To the best of our knowledge, ParaNames provides the broadest coverage of entities and languages available of any parallel name resource to date.
The release of this resource enables multifaceted research in names, including name translation, named entity recognition and entity linking, especially in less-resourced languages.
In addition to describing our process for creating this resource, we have demonstrated its usefulness two downstream tasks: canonical name translation and multilingual named entity recognition.
ParaNames can also be used as a building block for other benchmark datasets, such as named entity-centered evaluations of large language models.

\section{Ethics and Broader Impact}

We believe that the creation of this resource will benefit the speakers of the included languages by enabling improvements to language technology and access to information in more languages.
This resource consists only of information voluntarily provided to a user-edited database regarding notable entities, and does not include data collected from sources like social media that users did not know would become part of a public dataset.

However, like any language technology resource, this work could have unanticipated negative impact, and this impact could be magnified because some of this resource contains data in the languages of marginalized and minoritized populations.

A potential risk in using this resource is that quality issues in Wikidata can be passed to downstream systems, resulting in unexpectedly poor performance.
As an extreme example of this, much of the content of Scots Wikipedia and associated content in Wikidata was found to have been created or edited by someone with minimal proficiency in the language,\footnote{\href{https://www.theguardian.com/uk-news/2020/aug/26/shock-an-aw-us-teenager-wrote-huge-slice-of-scots-wikipedia}{Shock an aw: US teenager wrote huge slice of Scots Wikipedia, \emph{The Guardian}, August 26th 2020.}} and this data was used in the training of Multilingual BERT \citep{devlin-etal-2019-bert}.
We encourage users of this resource who build systems to collaborate with native speakers to verify data quality in the specific languages used.

\section{Bibliographical References}\label{sec:reference}
\bibliographystyle{lrec-coling2024-natbib}
\bibliography{anthology,custom}

\section{Language Resource References}
\label{lr:ref}
\bibliographystylelanguageresource{lrec-coling2024-natbib}
\bibliographylanguageresource{custom_languageresource}

\appendix

\section{Appendix}

\subsection{Canonical Name Translation}

For the full results table, see Table~\ref{bigtable-all-results}.

\subsection{Hyperparameters for NER}

\begin{table}[ht]
\small
\begin{tabular}{ll}
\toprule
Hyperparameter & Values \\
\midrule
Type disambiguation in resource & on, off \\
Autoencoder loss & on, off \\
Soft gaz. features & \\
\hspace{1em} \dots at CRF layer & on, off \\
\hspace{1em} \dots at LSTM layer & on, off \\
\midrule
LSTM embedding dimension & 128, 200, 256 \\
Random seed & \texttt{randint} \\
\bottomrule
\end{tabular}
\caption{Soft gazetteer feature configurations (top) and hyperparameters to average over (bottom). Setting all of the top values to \textit{off} corresponds to our baseline configuration.}
\label{tab:softgaz-ablation}
\end{table}

\begin{landscape}
\begin{table}
\centering
\resizebox{9in}{!}{

\begin{tabular}{lllllllllllllllllll}
\toprule
{} & \multicolumn{3}{c}{None} & \multicolumn{3}{c}{Script only} & \multicolumn{3}{c}{Language only} & \multicolumn{3}{c}{Language + script} & \multicolumn{3}{c}{Language + type} & \multicolumn{3}{c}{Language + type + script} \\
            \cmidrule(lr){2-4} \cmidrule(lr){5-7} \cmidrule(lr){8-10} \cmidrule(lr){11-13} \cmidrule(lr){14-16} \cmidrule(lr){17-19}
{} &          Accuracy &             CER &                F1 &          Accuracy &             CER &                F1 &          Accuracy &             CER &                F1 &          Accuracy &             CER &                F1 &          Accuracy &              CER &                F1 &                 Accuracy &             CER &                F1 \\
\midrule
Arabic     &  30.66 $\pm$ 0.11 &  0.25 $\pm$ 0.0 &  91.09 $\pm$ 0.02 &  30.62 $\pm$ 0.21 &  0.25 $\pm$ 0.0 &  91.08 $\pm$ 0.04 &  30.77 $\pm$ 0.13 &  0.25 $\pm$ 0.0 &  91.13 $\pm$ 0.02 &  30.91 $\pm$ 0.11 &  0.25 $\pm$ 0.0 &  91.11 $\pm$ 0.01 &  31.88 $\pm$ 0.12 &  0.24 $\pm$ 0.0 &  91.43 $\pm$ 0.01 &         31.72 $\pm$ 0.09 &  0.24 $\pm$ 0.0 &  91.39 $\pm$ 0.02 \\
Armenian   &   38.04 $\pm$ 0.1 &  0.25 $\pm$ 0.0 &  90.96 $\pm$ 0.01 &  37.96 $\pm$ 0.08 &  0.25 $\pm$ 0.0 &  90.94 $\pm$ 0.02 &  38.05 $\pm$ 0.09 &  0.25 $\pm$ 0.0 &  90.93 $\pm$ 0.02 &  38.16 $\pm$ 0.14 &  0.25 $\pm$ 0.0 &  90.96 $\pm$ 0.02 &  38.76 $\pm$ 0.16 &  0.25 $\pm$ 0.0 &  91.12 $\pm$ 0.02 &         39.04 $\pm$ 0.15 &  0.25 $\pm$ 0.0 &  91.11 $\pm$ 0.02 \\
Georgian   &   31.6 $\pm$ 0.16 &  0.28 $\pm$ 0.0 &  89.48 $\pm$ 0.01 &  31.54 $\pm$ 0.12 &  0.28 $\pm$ 0.0 &  89.48 $\pm$ 0.01 &  31.56 $\pm$ 0.12 &  0.28 $\pm$ 0.0 &  89.44 $\pm$ 0.02 &  31.59 $\pm$ 0.13 &  0.28 $\pm$ 0.0 &  89.47 $\pm$ 0.01 &  32.67 $\pm$ 0.12 &  0.28 $\pm$ 0.0 &  89.64 $\pm$ 0.02 &         32.62 $\pm$ 0.13 &  0.28 $\pm$ 0.0 &  89.64 $\pm$ 0.01 \\
Greek      &  31.12 $\pm$ 0.14 &  0.27 $\pm$ 0.0 &  90.05 $\pm$ 0.02 &  31.08 $\pm$ 0.04 &  0.27 $\pm$ 0.0 &  90.05 $\pm$ 0.02 &  31.04 $\pm$ 0.07 &  0.27 $\pm$ 0.0 &  90.05 $\pm$ 0.02 &  31.08 $\pm$ 0.16 &  0.27 $\pm$ 0.0 &  90.08 $\pm$ 0.02 &  31.67 $\pm$ 0.11 &  0.26 $\pm$ 0.0 &  90.17 $\pm$ 0.02 &         31.96 $\pm$ 0.06 &  0.26 $\pm$ 0.0 &  90.24 $\pm$ 0.02 \\
Hebrew     &  17.54 $\pm$ 0.07 &  0.33 $\pm$ 0.0 &  88.31 $\pm$ 0.05 &  17.64 $\pm$ 0.19 &  0.33 $\pm$ 0.0 &  88.35 $\pm$ 0.02 &  17.66 $\pm$ 0.12 &  0.33 $\pm$ 0.0 &  88.37 $\pm$ 0.02 &  17.83 $\pm$ 0.07 &  0.33 $\pm$ 0.0 &   88.4 $\pm$ 0.03 &  18.46 $\pm$ 0.15 &  0.32 $\pm$ 0.0 &  88.54 $\pm$ 0.03 &         18.34 $\pm$ 0.11 &  0.32 $\pm$ 0.0 &  88.57 $\pm$ 0.02 \\
Japanese   &   29.2 $\pm$ 0.04 &   0.3 $\pm$ 0.0 &   88.9 $\pm$ 0.02 &    29.6 $\pm$ 0.1 &   0.3 $\pm$ 0.0 &  88.91 $\pm$ 0.01 &  29.51 $\pm$ 0.09 &   0.3 $\pm$ 0.0 &  88.96 $\pm$ 0.04 &  29.56 $\pm$ 0.09 &   0.3 $\pm$ 0.0 &  88.98 $\pm$ 0.02 &  29.97 $\pm$ 0.13 &   0.3 $\pm$ 0.0 &  89.07 $\pm$ 0.03 &          29.6 $\pm$ 0.08 &   0.3 $\pm$ 0.0 &  89.04 $\pm$ 0.02 \\
Kazakh     &   47.44 $\pm$ 0.1 &  0.17 $\pm$ 0.0 &  93.33 $\pm$ 0.01 &  47.39 $\pm$ 0.09 &  0.17 $\pm$ 0.0 &  93.36 $\pm$ 0.01 &  48.36 $\pm$ 0.17 &  0.16 $\pm$ 0.0 &  93.54 $\pm$ 0.01 &  48.35 $\pm$ 0.14 &  0.16 $\pm$ 0.0 &  93.55 $\pm$ 0.02 &  49.14 $\pm$ 0.11 &  0.16 $\pm$ 0.0 &  93.62 $\pm$ 0.02 &          49.0 $\pm$ 0.18 &  0.16 $\pm$ 0.0 &  93.63 $\pm$ 0.02 \\
Korean     &  31.16 $\pm$ 0.07 &  0.29 $\pm$ 0.0 &  89.48 $\pm$ 0.02 &   31.1 $\pm$ 0.18 &  0.29 $\pm$ 0.0 &  89.47 $\pm$ 0.05 &   31.21 $\pm$ 0.1 &  0.29 $\pm$ 0.0 &  89.51 $\pm$ 0.04 &  30.89 $\pm$ 0.08 &  0.29 $\pm$ 0.0 &  89.51 $\pm$ 0.02 &  32.28 $\pm$ 0.19 &  0.28 $\pm$ 0.0 &  89.84 $\pm$ 0.03 &         32.34 $\pm$ 0.05 &  0.28 $\pm$ 0.0 &  89.82 $\pm$ 0.04 \\
Latvian    &  73.48 $\pm$ 0.04 &  0.11 $\pm$ 0.0 &  96.41 $\pm$ 0.02 &  73.84 $\pm$ 0.13 &   0.1 $\pm$ 0.0 &  96.45 $\pm$ 0.02 &  74.86 $\pm$ 0.16 &   0.1 $\pm$ 0.0 &  96.63 $\pm$ 0.02 &  74.66 $\pm$ 0.14 &   0.1 $\pm$ 0.0 &  96.66 $\pm$ 0.01 &  75.26 $\pm$ 0.03 &   0.1 $\pm$ 0.0 &   96.7 $\pm$ 0.01 &         75.34 $\pm$ 0.08 &   0.1 $\pm$ 0.0 &   96.7 $\pm$ 0.02 \\
Lithuanian &  79.66 $\pm$ 0.05 &  0.09 $\pm$ 0.0 &  96.44 $\pm$ 0.02 &   79.7 $\pm$ 0.06 &  0.09 $\pm$ 0.0 &  96.46 $\pm$ 0.01 &  80.01 $\pm$ 0.07 &  0.09 $\pm$ 0.0 &  96.53 $\pm$ 0.01 &  80.16 $\pm$ 0.06 &  0.09 $\pm$ 0.0 &  96.55 $\pm$ 0.01 &  80.56 $\pm$ 0.08 &  0.09 $\pm$ 0.0 &   96.6 $\pm$ 0.01 &         80.58 $\pm$ 0.11 &  0.09 $\pm$ 0.0 &   96.6 $\pm$ 0.01 \\
Persian    &  25.75 $\pm$ 0.22 &  0.28 $\pm$ 0.0 &  89.47 $\pm$ 0.03 &   25.48 $\pm$ 0.1 &  0.28 $\pm$ 0.0 &  89.43 $\pm$ 0.01 &   25.9 $\pm$ 0.12 &  0.28 $\pm$ 0.0 &  89.54 $\pm$ 0.03 &  25.68 $\pm$ 0.16 &  0.28 $\pm$ 0.0 &   89.5 $\pm$ 0.02 &  26.92 $\pm$ 0.12 &  0.27 $\pm$ 0.0 &  89.88 $\pm$ 0.01 &          26.6 $\pm$ 0.16 &  0.27 $\pm$ 0.0 &  89.85 $\pm$ 0.01 \\
Russian    &  45.16 $\pm$ 0.08 &  0.23 $\pm$ 0.0 &  91.92 $\pm$ 0.02 &  45.02 $\pm$ 0.07 &  0.23 $\pm$ 0.0 &  91.92 $\pm$ 0.03 &  45.18 $\pm$ 0.04 &  0.23 $\pm$ 0.0 &  91.95 $\pm$ 0.03 &   45.5 $\pm$ 0.11 &  0.22 $\pm$ 0.0 &  92.01 $\pm$ 0.03 &  45.65 $\pm$ 0.06 &  0.22 $\pm$ 0.0 &  92.05 $\pm$ 0.02 &         45.78 $\pm$ 0.04 &  0.22 $\pm$ 0.0 &  92.07 $\pm$ 0.02 \\
Swedish    &  90.14 $\pm$ 0.06 &  0.05 $\pm$ 0.0 &  98.11 $\pm$ 0.01 &  90.11 $\pm$ 0.04 &  0.05 $\pm$ 0.0 &   98.12 $\pm$ 0.0 &  90.26 $\pm$ 0.12 &  0.05 $\pm$ 0.0 &  98.14 $\pm$ 0.01 &   90.16 $\pm$ 0.1 &  0.05 $\pm$ 0.0 &  98.11 $\pm$ 0.01 &  90.34 $\pm$ 0.08 &  0.05 $\pm$ 0.0 &  98.14 $\pm$ 0.02 &         90.31 $\pm$ 0.13 &  0.05 $\pm$ 0.0 &  98.13 $\pm$ 0.02 \\
Tajik      &  48.38 $\pm$ 0.18 &  0.22 $\pm$ 0.0 &  92.27 $\pm$ 0.03 &  48.53 $\pm$ 0.13 &  0.22 $\pm$ 0.0 &  92.21 $\pm$ 0.02 &  51.25 $\pm$ 0.15 &  0.21 $\pm$ 0.0 &  92.51 $\pm$ 0.02 &   51.24 $\pm$ 0.1 &  0.21 $\pm$ 0.0 &  92.49 $\pm$ 0.01 &  51.56 $\pm$ 0.21 &  0.21 $\pm$ 0.0 &  92.55 $\pm$ 0.06 &         51.55 $\pm$ 0.15 &  0.21 $\pm$ 0.0 &  92.54 $\pm$ 0.03 \\
Thai       &    38.1 $\pm$ 0.2 &  0.27 $\pm$ 0.0 &  90.53 $\pm$ 0.01 &  38.24 $\pm$ 0.28 &  0.27 $\pm$ 0.0 &  90.46 $\pm$ 0.03 &  38.55 $\pm$ 0.11 &  0.27 $\pm$ 0.0 &  90.53 $\pm$ 0.03 &  38.67 $\pm$ 0.12 &  0.27 $\pm$ 0.0 &  90.55 $\pm$ 0.03 &  39.59 $\pm$ 0.22 &  0.26 $\pm$ 0.0 &  90.74 $\pm$ 0.04 &          39.63 $\pm$ 0.1 &  0.27 $\pm$ 0.0 &  90.72 $\pm$ 0.01 \\
Urdu       &   24.42 $\pm$ 0.1 &  0.27 $\pm$ 0.0 &  90.19 $\pm$ 0.01 &   24.49 $\pm$ 0.1 &  0.27 $\pm$ 0.0 &   90.2 $\pm$ 0.04 &  26.18 $\pm$ 0.12 &  0.25 $\pm$ 0.0 &   90.6 $\pm$ 0.03 &  26.21 $\pm$ 0.13 &  0.25 $\pm$ 0.0 &  90.58 $\pm$ 0.01 &  27.02 $\pm$ 0.19 &  0.25 $\pm$ 0.0 &  90.81 $\pm$ 0.03 &         27.12 $\pm$ 0.11 &  0.25 $\pm$ 0.0 &  90.82 $\pm$ 0.02 \\
Vietnamese &  86.48 $\pm$ 0.03 &   0.1 $\pm$ 0.0 &   96.73 $\pm$ 0.0 &  86.46 $\pm$ 0.02 &   0.1 $\pm$ 0.0 &  96.71 $\pm$ 0.01 &  86.78 $\pm$ 0.07 &   0.1 $\pm$ 0.0 &  96.76 $\pm$ 0.01 &   86.9 $\pm$ 0.06 &   0.1 $\pm$ 0.0 &  96.75 $\pm$ 0.02 &  87.02 $\pm$ 0.07 &  0.09 $\pm$ 0.0 &  96.82 $\pm$ 0.02 &          86.94 $\pm$ 0.1 &  0.09 $\pm$ 0.0 &  96.79 $\pm$ 0.02 \\
\midrule
Overall    &   45.2 $\pm$ 0.05 &  0.22 $\pm$ 0.0 &   91.98 $\pm$ 0.0 &  45.22 $\pm$ 0.06 &  0.22 $\pm$ 0.0 &  91.98 $\pm$ 0.01 &  45.71 $\pm$ 0.06 &  0.22 $\pm$ 0.0 &  92.07 $\pm$ 0.01 &  45.74 $\pm$ 0.02 &  0.22 $\pm$ 0.0 &   92.07 $\pm$ 0.0 &   46.4 $\pm$ 0.05 &  0.22 $\pm$ 0.0 &  92.22 $\pm$ 0.01 &         46.38 $\pm$ 0.04 &  0.22 $\pm$ 0.0 &  92.22 $\pm$ 0.01 \\
\midrule
Arabic     &                 - &               - &                 - &                 - &               - &                 - &  46.54 $\pm$ 0.09 &  0.19 $\pm$ 0.0 &  93.04 $\pm$ 0.02 &                 - &               - &                 - &  46.77 $\pm$ 0.13 &  0.19 $\pm$ 0.0 &  93.09 $\pm$ 0.01 &                        - &               - &                 - \\
Armenian   &                 - &               - &                 - &                 - &               - &                 - &  47.42 $\pm$ 0.05 &  0.22 $\pm$ 0.0 &  92.84 $\pm$ 0.01 &                 - &               - &                 - &  47.95 $\pm$ 0.08 &  0.22 $\pm$ 0.0 &   92.9 $\pm$ 0.03 &                        - &               - &                 - \\
Georgian   &                 - &               - &                 - &                 - &               - &                 - &  48.94 $\pm$ 0.16 &  0.21 $\pm$ 0.0 &  92.81 $\pm$ 0.01 &                 - &               - &                 - &   51.0 $\pm$ 0.23 &   0.2 $\pm$ 0.0 &  93.08 $\pm$ 0.03 &                        - &               - &                 - \\
Greek      &                 - &               - &                 - &                 - &               - &                 - &  30.77 $\pm$ 0.07 &  0.29 $\pm$ 0.0 &  89.78 $\pm$ 0.05 &                 - &               - &                 - &  31.22 $\pm$ 0.15 &  0.29 $\pm$ 0.0 &  89.86 $\pm$ 0.04 &                        - &               - &                 - \\
Hebrew     &                 - &               - &                 - &                 - &               - &                 - &  37.32 $\pm$ 0.07 &  0.25 $\pm$ 0.0 &  91.14 $\pm$ 0.02 &                 - &               - &                 - &  37.83 $\pm$ 0.06 &  0.25 $\pm$ 0.0 &  91.22 $\pm$ 0.03 &                        - &               - &                 - \\
Japanese   &                 - &               - &                 - &                 - &               - &                 - &  26.95 $\pm$ 0.08 &  0.37 $\pm$ 0.0 &  85.49 $\pm$ 0.04 &                 - &               - &                 - &   27.3 $\pm$ 0.09 &  0.36 $\pm$ 0.0 &  85.65 $\pm$ 0.06 &                        - &               - &                 - \\
Kazakh     &                 - &               - &                 - &                 - &               - &                 - &   58.0 $\pm$ 0.11 &  0.16 $\pm$ 0.0 &  94.48 $\pm$ 0.01 &                 - &               - &                 - &    58.3 $\pm$ 0.1 &  0.16 $\pm$ 0.0 &  94.58 $\pm$ 0.02 &                        - &               - &                 - \\
Korean     &                 - &               - &                 - &                 - &               - &                 - &   42.32 $\pm$ 0.1 &  0.31 $\pm$ 0.0 &  88.74 $\pm$ 0.02 &                 - &               - &                 - &  42.57 $\pm$ 0.09 &  0.31 $\pm$ 0.0 &  88.85 $\pm$ 0.03 &                        - &               - &                 - \\
Latvian    &                 - &               - &                 - &                 - &               - &                 - &  73.05 $\pm$ 0.25 &  0.11 $\pm$ 0.0 &  96.17 $\pm$ 0.01 &                 - &               - &                 - &  73.81 $\pm$ 0.19 &  0.11 $\pm$ 0.0 &  96.26 $\pm$ 0.02 &                        - &               - &                 - \\
Lithuanian &                 - &               - &                 - &                 - &               - &                 - &  78.73 $\pm$ 0.44 &  0.09 $\pm$ 0.0 &  96.55 $\pm$ 0.04 &                 - &               - &                 - &   79.3 $\pm$ 0.31 &  0.09 $\pm$ 0.0 &  96.74 $\pm$ 0.03 &                        - &               - &                 - \\
Persian    &                 - &               - &                 - &                 - &               - &                 - &   41.9 $\pm$ 0.13 &  0.24 $\pm$ 0.0 &  90.59 $\pm$ 0.04 &                 - &               - &                 - &   42.1 $\pm$ 0.21 &  0.24 $\pm$ 0.0 &  90.64 $\pm$ 0.05 &                        - &               - &                 - \\
Russian    &                 - &               - &                 - &                 - &               - &                 - &  42.42 $\pm$ 0.12 &  0.35 $\pm$ 0.0 &  89.03 $\pm$ 0.04 &                 - &               - &                 - &  43.26 $\pm$ 0.14 &  0.35 $\pm$ 0.0 &  89.27 $\pm$ 0.05 &                        - &               - &                 - \\
Swedish    &                 - &               - &                 - &                 - &               - &                 - &   87.96 $\pm$ 0.2 &  0.06 $\pm$ 0.0 &  97.74 $\pm$ 0.03 &                 - &               - &                 - &  88.31 $\pm$ 0.06 &  0.06 $\pm$ 0.0 &  97.82 $\pm$ 0.01 &                        - &               - &                 - \\
Tajik      &                 - &               - &                 - &                 - &               - &                 - &    56.3 $\pm$ 0.1 &  0.17 $\pm$ 0.0 &  94.04 $\pm$ 0.06 &                 - &               - &                 - &  56.82 $\pm$ 0.11 &  0.17 $\pm$ 0.0 &  94.07 $\pm$ 0.07 &                        - &               - &                 - \\
Thai       &                 - &               - &                 - &                 - &               - &                 - &  14.59 $\pm$ 0.13 &   0.4 $\pm$ 0.0 &  84.71 $\pm$ 0.07 &                 - &               - &                 - &  15.02 $\pm$ 0.04 &  0.39 $\pm$ 0.0 &  85.26 $\pm$ 0.05 &                        - &               - &                 - \\
Urdu       &                 - &               - &                 - &                 - &               - &                 - &   18.03 $\pm$ 0.2 &  0.38 $\pm$ 0.0 &  84.17 $\pm$ 0.04 &                 - &               - &                 - &  17.96 $\pm$ 0.16 &  0.38 $\pm$ 0.0 &   84.3 $\pm$ 0.05 &                        - &               - &                 - \\
Vietnamese &                 - &               - &                 - &                 - &               - &                 - &  77.71 $\pm$ 0.28 &  0.16 $\pm$ 0.0 &  93.67 $\pm$ 0.04 &                 - &               - &                 - &  78.17 $\pm$ 0.16 &  0.16 $\pm$ 0.0 &  93.73 $\pm$ 0.04 &                        - &               - &                 - \\
\midrule
Overall    &                 - &               - &                 - &                 - &               - &                 - &  48.76 $\pm$ 0.05 &  0.23 $\pm$ 0.0 &  91.47 $\pm$ 0.01 &                 - &               - &                 - &  49.27 $\pm$ 0.04 &  0.23 $\pm$ 0.0 &  91.61 $\pm$ 0.02 &                        - &               - &                 - \\
\bottomrule
\end{tabular}

}
\caption{Canonical name translation results for the X $\rightarrow$ English (top) and English $\rightarrow$ X directions across different special token settings.}
\label{bigtable-all-results}
\end{table}
\end{landscape}

\end{document}